\documentclass[conference]{IEEEtran}
\IEEEoverridecommandlockouts
\usepackage{amsmath,amssymb,amsfonts}
\usepackage{algorithmic}
\usepackage{graphicx}
\usepackage{textcomp}
\usepackage{epsfig}
\usepackage[dvipsnames]{xcolor}
\usepackage{multirow}
\usepackage{multicol}
\usepackage{subcaption}
\usepackage{arydshln}
\usepackage{booktabs}
\usepackage{bbding}
\usepackage{tablefootnote}
\usepackage{threeparttable}
\usepackage{pifont}
\usepackage[accsupp]{axessibility}
\usepackage{caption}
\usepackage[hidelinks, colorlinks=true, linkcolor=magenta, urlcolor=magenta]{hyperref}
\usepackage[capitalize]{cleveref}

\def\BibTeX{{\rm B\kern-.05em{\sc i\kern-.025em b}\kern-.08em
    T\kern-.1667em\lower.7ex\hbox{E}\kern-.125emX}}
\begin{document}

\title{DPCD: A Quality Assessment Database for Dynamic Point Clouds}

\author{\IEEEauthorblockN{Yating Liu\textsuperscript{1}, Yujie Zhang\textsuperscript{1}, Qi Yang\textsuperscript{2}, Yiling Xu\textsuperscript{1$^{\dagger}$}\thanks{\textsuperscript{$^{\dagger}$} Corresponding author.}, Zhu Li\textsuperscript{2}, Ye-Kui Wang\textsuperscript{3}}

\IEEEauthorblockA{\textsuperscript{1}\textit{Shanghai Jiao Tong University}, \textsuperscript{2}\textit{University of Missouri-Kansas City}, \textsuperscript{3}\textit{Bytedance}}
\IEEEauthorblockA{\textsuperscript{1}\{Olivialyt, yujie19981026, yl.xu\}@sjtu.edu.cn, \textsuperscript{2}\{qiyang, lizhu\}@umkc.edu, \textsuperscript{3}\{yekui.wang\}@bytedance.com}}


\maketitle
\begin{abstract}
Recently, the advancements in Virtual/Augmented Reality (VR/AR) have driven the demand for Dynamic Point Clouds (DPC).  Unlike static point clouds, DPCs are capable of capturing temporal changes within objects or scenes, offering a more accurate simulation of the real world. While significant progress has been made in the quality assessment research of static point cloud, little study has been done on Dynamic Point Cloud Quality Assessment (DPCQA), which hinders the development of quality-oriented applications, such as interframe compression and transmission in practical scenarios. In this paper, we introduce a large-scale DPCQA database, named DPCD, which includes 15 reference DPCs and 525 distorted DPCs from seven types of lossy compression and noise distortion. By rendering these samples to Processed Video Sequences (PVS), a comprehensive subjective experiment is conducted to obtain Mean Opinion Scores (MOS) from 21 viewers for analysis. The characteristic of contents, impact of various distortions, and accuracy of MOSs are presented to validate the heterogeneity and reliability of the proposed database. Furthermore, we evaluate the performance of several objective metrics on DPCD. The experiment results show that DPCQA is more challenge than that of static point cloud. The DPCD, which serves as a catalyst for new research endeavors on DPCQA, is publicly available at \href{https://huggingface.co/datasets/Olivialyt/DPCD}{https://huggingface.co/datasets/Olivialyt/DPCD}.

\end{abstract}
\section{Introduction}
\label{sec:intro}
Point Clouds (PC), as one of the most representative forms of data in immersive media, are experiencing increasing demand in various fields, such as autonomous driving~\cite{cui2021deep} and medical imaging~\cite{cheng2020morphing}.
A point cloud consists of a collection of discrete points, each described by its coordinates in 3D space, along with additional attributes such as color and normal vectors. Given the inevitable distortions introduced to point clouds in practical applications, which impact the perceptual quality, research on PCQA has become a hotspot. PCQA can be roughly classified into subjective and objective quality assessment. Subjective quality assessment is considered the most reliable method, involving the invitation of viewers to evaluate the quality of distorted point clouds in a controlled testing environment. Objective quality assessment explores metrics that are strongly correlated with human perceptual quality, aiming to replace subjective evaluations in practical applications and thereby reduce time and costs.

\begin{table}[t]\small
\centering
\renewcommand{\arraystretch}{1.1}
\setlength{\tabcolsep}{6pt}
\caption{PCQA database survey.}
\begin{tabular}{cccc}
\hline
\textbf{Type} & \textbf{Name} & \textbf{Scale} & \textbf{Distortion Types} \\
\hline
\noalign{\vskip 0.3em}
\multirow{4}{*}{Static} 
 &SJTU-PCQA~\cite{yang2020predicting} & 420 & 7 \\
 &WPC~\cite{liu2022perceptual} & 740 & 5 \\
 &LS-PCQA~\cite{liu2023point} & 930 & 31 \\
 &BASICS~\cite{ak2024basics} & 1494 & 4 
\\
 \hdashline
\noalign{\vskip 0.3em}
\multirow{3}{*}{Dynamic} 
&vsenseVVDB~\cite{zerman2019subjective} & 32 & 1 \\
&vsenseVVDB2~\cite{zerman2020textured} & 128 & 3 \\
&DPCD (Proposed) & 525 & 7 \\
\hline
\end{tabular}
\label{tab:dataset_all}
\end{table}

In recent years, advancements in 3D acquisition devices have made VR and AR more accessible than ever. To provide users with more interactive and immersive experiences, DPC has gained significant attention. Unlike static point clouds, DPCs incorporate a temporal dimension, which enables a more realistic representation of 3D environments, simulating the dynamic nature of the real world. However, due to the large volume of data they contain, DPCs require more efficient compression and transmission techniques before practical utilizations. Similar to static point clouds, these processes incur distortion and impact perceived quality. Consequently, Dynamic Point Cloud Quality Assessment (DPCQA) has become an increasingly important research focus in both industry and academia.

Currently, significant progress has been made in Static Point Cloud Quality Assessment (SPCQA), but research on DPCQA remains limited. For comparison, we list existing PCQA databases in~\Cref{tab:dataset_all}. Previous studies have conducted DPCQA evaluation by proposing new benchmarks. For example, vsenseVVDB~\cite{zerman2019subjective} and vsenseVVDB2~\cite{zerman2020textured} investigate the impact of compression on point clouds. However, these databases have two main drawbacks. \textbf{1) Limited Scale.} Compared to SPCQA databases, existing DPCQA databases are typically small in scale, regardless of reference or distorted samples. \textbf{2) Lack of Distortion Types.} These databases focus solely on traditional compression algorithms, overlooking emerging learning-based compression techniques and distortions from other scenarios. The above weaknesses limit the generalizability of these databases, and also hinder the development and validation of objective DPCQA metrics. Especially, the Call for Proposals (CfP) for learning-based DPC compression technology within Moving Picture Experts Group (MPEG) - WG 2~\cite{MPEG-DPC-Coding} highlights the need for reliable objective DPCQA metrics. Besides, although many high-performing objective SPCQA metrics have been developed, whether they are suitable for DPCs is uncertain. 

In view of the above challenges, to effectively promote the development of DPCQA and relevant algorithms such as compression and transmission of DPCs, we create a large-scale DPCQA database named DPCD, which contains rich contents and multiple types of distortion. 15 high-quality reference DPC sequences are selected and seven types of distortion are injected at different levels, resulting in a total of 525 distorted DPCs. To conduct subjective experiments, all samples are rendered into Processed Video Sequences (PVS) and participants are invited to score them in a lab environment to collect MOSs. 
The diversity of source content, the accuracy of the MOSs, and the influence of different types of distortion are demonstrated. Finally, we evaluate the performance of multiple objective metrics and conduct detailed analysis of results to provide useful insight for future DPCQA research.

\begin{figure}[t]
\centering
\includegraphics[width=0.49\textwidth]{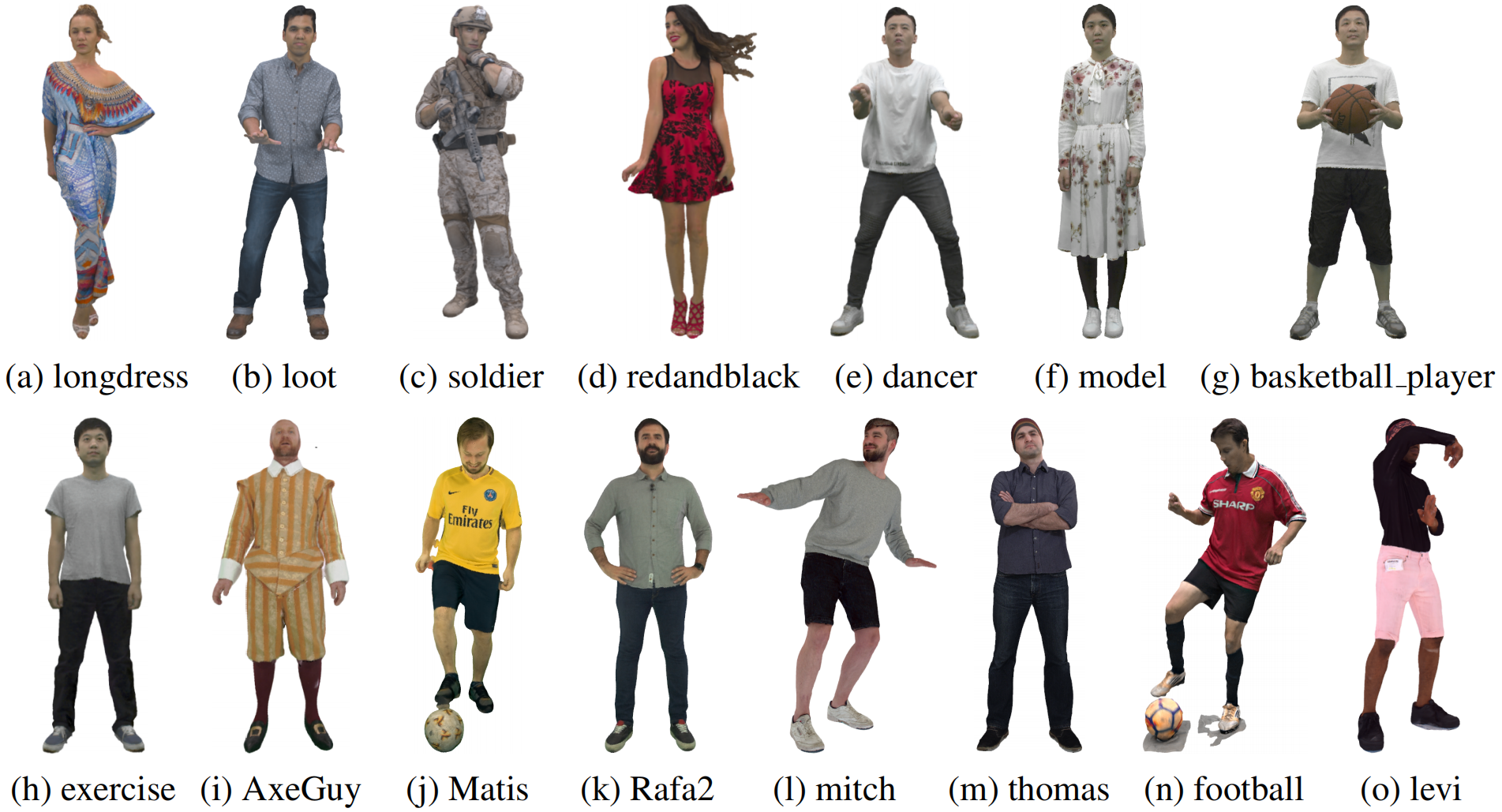}
\caption{The projection of reference samples in DPCD.}
\label{fig:15_projections}
\end{figure}

\section{Database Construction}
\label{sec:Construction}

\subsection{Reference Selection and Pre-processing}
\noindent\textbf{Reference Selection.} 
Given that the primary application of DPCs is social communication within extended reality, we choose human DPCs for this study. To effectively advance the development of standard compression algorithms, we utilize DPC sequences~\cite{8i,owlii} provided by the MPEG and convert some dynamic meshes~\cite{xdprod,vsense} from MPEG into point clouds using grid sampling with a resolution of 1024, as proposed in~\cite{MPEG-MESH-metric}.
Moreover, we select some DPC sequences from other DPCQA database~\cite{zerman2020textured}. 

In total, 15 human DPCs are chosen as reference samples. 
Specifically, ``longdress", ``loot", ``soldier", and ``redandblack" are from 8i Voxelized Full Body Dataset~\cite{8i}, ``dancer", ``model", ``basketball-player", and ``exercise" are from Owlii Dataset~\cite{owlii}, ``AxeGuy", ``Matis", and ``Rafa2" are from vsenseVVDB2~\cite{zerman2020textured}, ``mitch", ``thomas", and ``football" are from~\cite{xdprod}, and ``levi" is from~\cite{vsense}. 
~\Cref{fig:15_projections} shows snapshots of all the reference DPCs, we list the number of points in the first frame of each reference sample in~\Cref{tab:database_points}.

\noindent\textbf{Pre-processing.} 
To ensure consistency in format and eliminate any potential factors that may introduce distortion, the DPCs are preprocessed.
For the four DPCs in the Owlii Dataset, each sequence contains 600 frames, from which we select the first 300 frames. For Levi with only 150 frames, we index the reverse order of the original 150 frames as frames 151 through 300 as proposed in \cite{TDMD}. As a result, all processed DPC sequences consist of 300 frames. Additionally, all samples are converted to the UTF-8 encoding format.

\subsection{Distortion Generation}
To investigate the impact of typical distortion types on perceptual quality during application, we consider seven distortion types.
First, we include traditional compression algorithms, as they are standardized by MPEG. Specifically, we choose two patterns of Geometry-based Point Cloud Compression (G-PCC)~\cite{mekuria2016design} and one pattern of Video-based Point Cloud Compression (V-PCC)~\cite{mekuria2016design}. Additionally, we select D-DPCC~\cite{fan2022d}, a learning-based DPC compression method. To simulate distortions arising from factors such as acquisition noise and resampling, we select Color Noise (CN), Down-Sampling (DS), and Geometry Gaussian Noise (GGN). For each distortion type, we apply three to six different levels, with processing details as follows:

\begin{table}[t]\small
\centering
\small
\renewcommand{\arraystretch}{1.1}
\setlength{\tabcolsep}{6pt}
\caption{Point count in the first frame of the references.}
\label{tab:database_points}
\begin{tabular}{l r | l r}
\toprule
\textbf{Name} & \textbf{Point Count} & \textbf{Name} & \textbf{Point Count} \\ 
\midrule
longdress & 765,821 & AxeGuy & 405,328 \\
loot & 784,142 & Matis & 405,696 \\
soldier & 1,059,810 & Rafa2 & 406,351 \\
redandblack & 729,133 & mitch & 1,593,174 \\
dancer & 2,025,053 & thomas & 1,288,784 \\
model & 1,866,582 & football & 1,355,369 \\
basketball-player & 2,201,357 & levi & 762,846 \\
exercise & 1,919,925 & --- & --- \\ 
\bottomrule
\end{tabular}
\end{table}

$\bullet$\textbf{G-PCC}: 
G-PCC encode point clouds in 3D space using octree or trisoup (triangle soup) methods. Color attributes can be encoded using either Region Adaptive Hierarchical Transform (RAHT) or Predicting/Lifting (PredLift) transform. We employ Octree-RAHT and Trisoup-RAHT for lossy compression. Octree-RAHT applies six distortion levels with Quantization Parameters (QP) of 51, 46, 40, 34, 28, and 22, while Trisoup-RAHT uses four levels with QPs of 40, 34, 28, and 22.

$\bullet$\textbf{V-PCC-C2RA}: The V-PCC using Category 2 Random Access (C2RA) mode is applied with five distortion levels, with QPs and occupancy map precision set as described in~\cite{mekuria2016design}. Both geometric and color attributes are lossily compressed. 

$\bullet$\textbf{D-DPCC}: D-DPCC utilizes sparse convolution for compression. Following previous studies~\cite{li2021quality,valenzise2018quality}, we adjust the Lagrange multiplier ($\lambda$) to control bitrate. To avoid information leakage, models trained on 8i Dataset are tested on other samples, while those trained on Owlii Dataset are tested on 8i Dataset. Three distortion levels are set with $\lambda$ of 0.1, 1, 10.

$\bullet$\textbf{CN}: Color noise affects the RGB values of points. We randomly modifies the RGB values of each point according to varying probabilities. Specifically, we randomly select 10$\%$, 30$\%$, 40$\%$, 50$\%$, 60$\%$, and 70$\%$ of points in each frame, and noise values of $\pm$10, $\pm$30, $\pm$40, $\pm$50, $\pm$60, $\pm$70 are added equally across the RGB channels. For example, for the first distortion level, we randomly select 10$\%$ points and modify their RGB values by a value within $\pm$10. If the modified color value (denoted as $c$) exceeds the valid range, we apply clipping: if $c<0$, set $c=0$; if $c>255$, set $c=255$.

$\bullet$\textbf{DS}: Down-sampling is a simple yet effective method to reduce data complexity. We use the Matlab function pcdownsample() to apply six distortion levels, with sampling rates set to 0.85, 0.7, 0.55, 0.4, 0.25, and 0.1.

$\bullet$\textbf{GGN}: GGN applys a geometry shift to each point in the point cloud. We set six levels, using random gaussian distribution with a mean of 0 and standard deviations of 0.05$\%$, 0.1$\%$, 0.2$\%$, 0.5$\%$, 0.7$\%$, and 1.2$\%$ of the Bounding Box (BB) coordinates.

\begin{figure}[t]
\captionsetup{font=small}
    \centering
    \begin{minipage}{0.49\textwidth} 
    \begin{subfigure}[b]{0.24\textwidth}
        \centering
        \includegraphics[height=1.9cm]{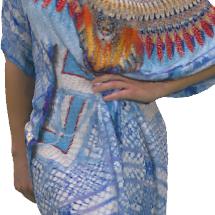}
        \caption{Reference}
    \end{subfigure}
    \begin{subfigure}[b]{0.24\textwidth}
        \centering
        \includegraphics[height=1.9cm]{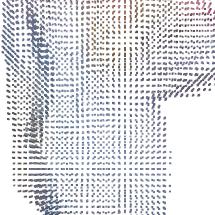}
        \caption{Octree-RAHT}
    \end{subfigure}
        \begin{subfigure}[b]{0.24\textwidth}
        \centering
        \includegraphics[height=1.9cm]{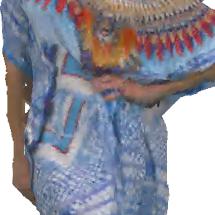}
        \caption{Trisoup-RAHT}
    \end{subfigure}
        \begin{subfigure}[b]{0.24\textwidth}
        \centering
        \includegraphics[height=1.9cm]{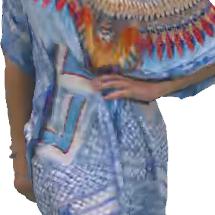}
        \caption{V-PCC-C2RA}
    \end{subfigure}
    \vspace{0.5em}
    \begin{subfigure}[b]{0.24\textwidth}
        \centering
        \includegraphics[height=1.9cm]{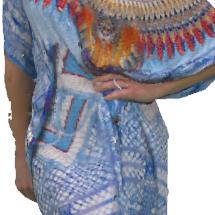}
        \caption{D-DPCC}
    \end{subfigure}
    \begin{subfigure}[b]{0.24\textwidth}
        \centering
        \includegraphics[height=1.9cm]{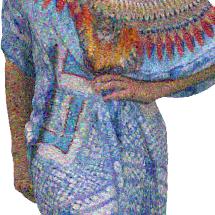}
        \caption{CN}
    \end{subfigure}
        \begin{subfigure}[b]{0.24\textwidth}
        \centering
        \includegraphics[height=1.9cm]{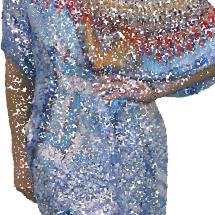}
        \caption{DS}
    \end{subfigure}
        \begin{subfigure}[b]{0.24\textwidth}
        \centering
        \includegraphics[height=1.9cm]{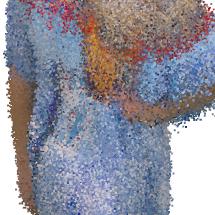}
        \caption{GGN}
    \end{subfigure}
\captionsetup{font=normal}
    \caption{The visual effects caused by different distortion types.}
    \label{fig:distortion_visual}
\end{minipage}
\end{figure}

We present local results for each distortion type at the maximum distortion level in~\Cref{fig:distortion_visual}, which indicates that the visual effects caused by different distortion types are distinct.

\subsection{PVS Generation}
DPC sequences can be rendered as 2D videos or presented immersively in 3D scenes using VR devices. While ITU-R BT.500~\cite{BT500} and ITU-T P.910~\cite{11scale-rating} provide detailed guidelines for video-based methods, no authoritative standards exist for VR-based subjective experiments, and the interactive nature of VR introduces variability in participants' viewing experiences. Therefore, we adopt a video-based approach.

With regard to the camera path, research~\cite{yang2023exploring} shows that participants focus mainly on human faces, with camera path variations having minimal impact as long as facial features are clearly visible. Therefore, we use a simple rendering method, fixing the camera distance and orienting it toward the front of the human model (\Cref{fig:15_projections}). DPCs are rendered into 1024×1024 images using Open3D and converted into a smooth PVS at 30 fps with FFMPEG’s libx264 standard.

\subsection{Subjective Experiment}
\subsubsection{Training and Test Session}
\label{sec:Training and Test Session}
To ensure the reliability of the collected subjective scores, we divide the database into two parts: training set and test set. The training set consists of 14 samples, which vary in point cloud content, distortion types and levels. The test set, including remaining 511 samples, is further divided into 7 subgroups of 73 samples to avoid the effects of visual fatigue. 

The subjective evaluation is carried out using the Double Stimulus Impairment Scale (DSIS), and the 11-grade voting scale proposed by ITU-T P.910~\cite{11scale-rating} is used. The experiments are conducted on an AOC U27N3G6R4B monitor with resolution of $3840 \times 2160$, in an indoor environment under normal lighting conditions. During the test session, each distorted sample is displayed for 10 seconds, followed by 5 seconds for scoring. For participants involved in multiple testing units, sufficient rest time is provided between adjacent units. 

A total of 26 participants are recruited for the subjective experiment. For each basic test unit, scores are collected from 22 different participants.

\subsubsection{Outlier Removal}
After collecting the subjective scores, we filter outliers from the raw scores to ensure data accuracy and reliability. Specifically, two consecutive steps are used. In the first step, we exclude outliers based on two ``trap" samples in the test set \cite{TSMD}. First, we randomly select one sample from each subgroup to repeat, ensuring that the two PVSs are not played consecutively. Secondly, we insert one sample of extremely poor quality into each subgroup. If the score difference of the duplicated PVSs or the score of the very low quality PVS is higher than 2, the scores collected from this viewer are considered incorrect. In the second step, we apply the outlier detection method recommended in ITU-R BT.500~\cite{BT500}. As a result, five viewers are identified and removed from the subjective scores.

\begin{figure}[t]
\centering
\begin{minipage}{0.49\textwidth} 
    \centering
    \begin{subfigure}[b]{0.48\textwidth}
    \centering 
        \includegraphics[height=3.6cm]{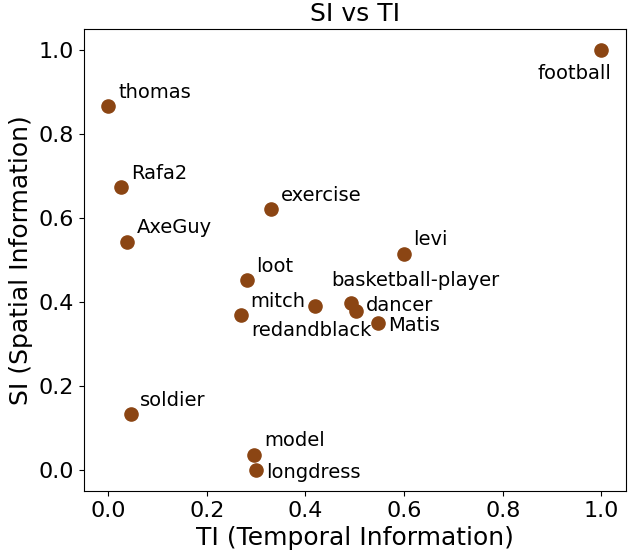}
        \caption{SI vs. TI}
        \label{fig:tisi}
    \end{subfigure}
    \begin{subfigure}[b]{0.48\textwidth}
    \centering
        \includegraphics[height=3.6cm]{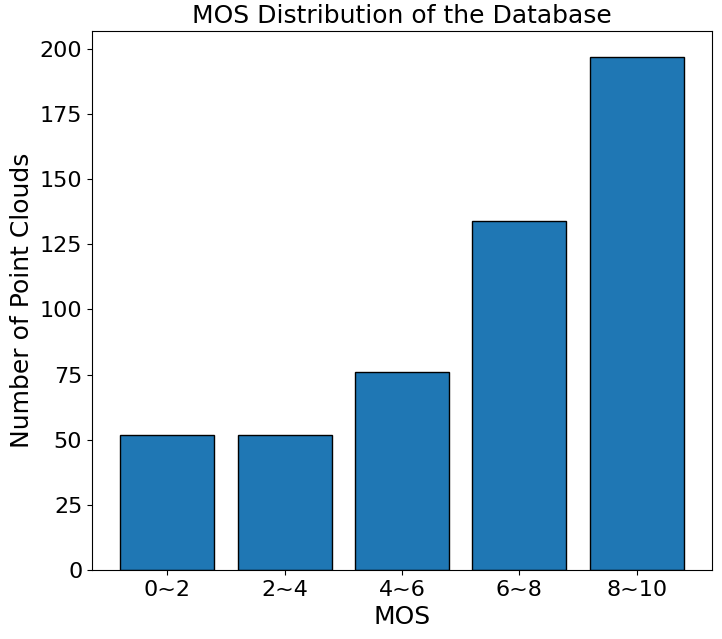}
        \caption{MOS Distribution}
        \label{fig:distribution}
    \end{subfigure}
\caption{Diversity of the source content and MOS distribution.}
\end{minipage}
\end{figure}

\section{Database Validation}
\subsection{Diversity of Content}
To validate the diversity of content, we calculate spatial perceptual information (SI) and temporal perceptual information (TI)~\cite{installations1999subjective} to measure geometry and dynamism complexities, respectively. More specifically, 15 reference PVSs are used to calculate SI and TI, and the results are shown in~\Cref{fig:tisi}. The relatively uniform distribution of the scatter points indicates the diversity of the source content in DPCD.

\begin{figure*}[t]
    \centering
    \begin{subfigure}[b]{0.21\textwidth}
        \centering
        \includegraphics[height=2.9cm]{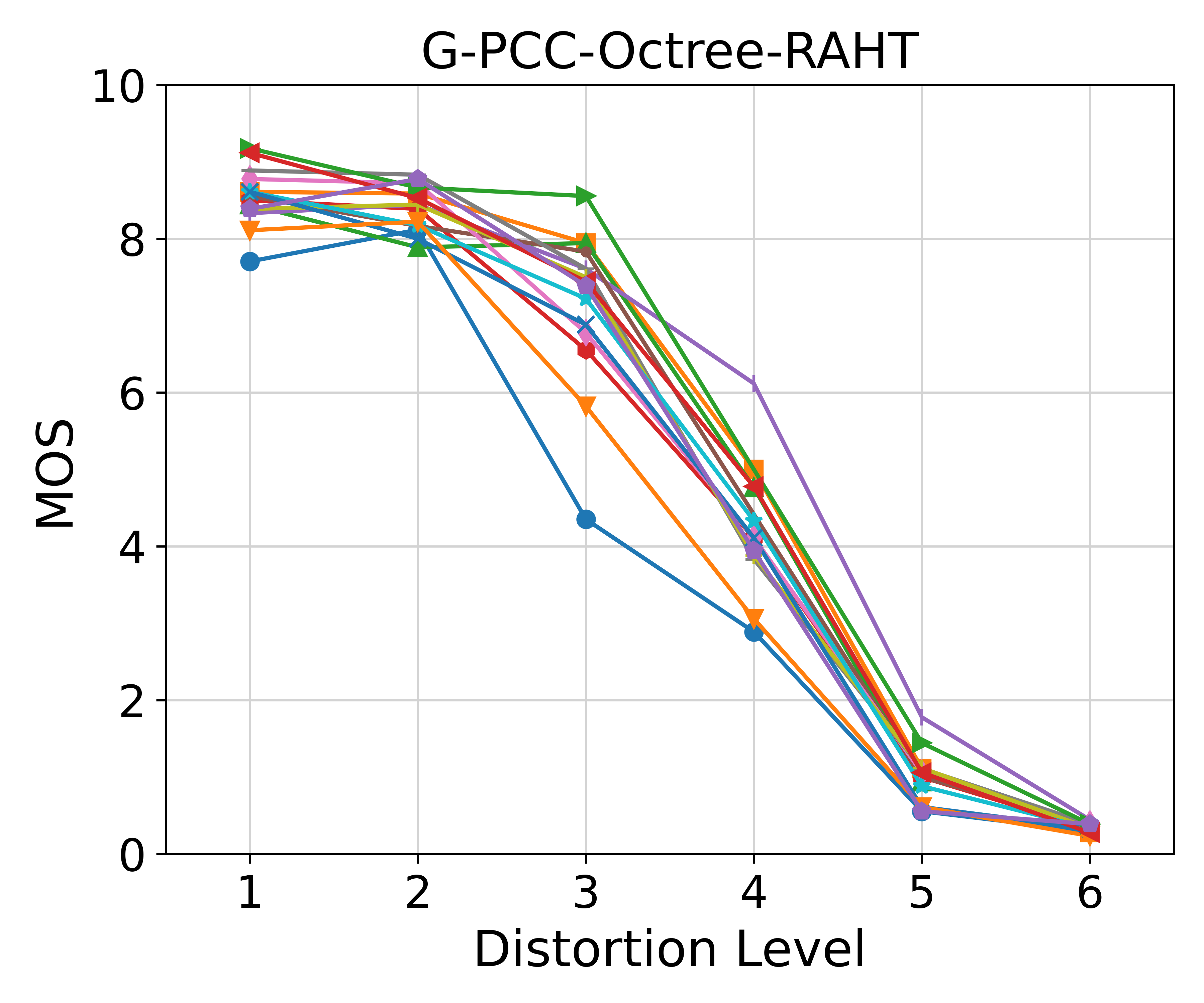}
        \caption{G-PCC-Octree-RAHT}
        \label{fig:G-PCC-Octree-RAHT}
    \end{subfigure}
    \begin{subfigure}[b]{0.21\textwidth}
        \centering
        \includegraphics[height=2.9cm]{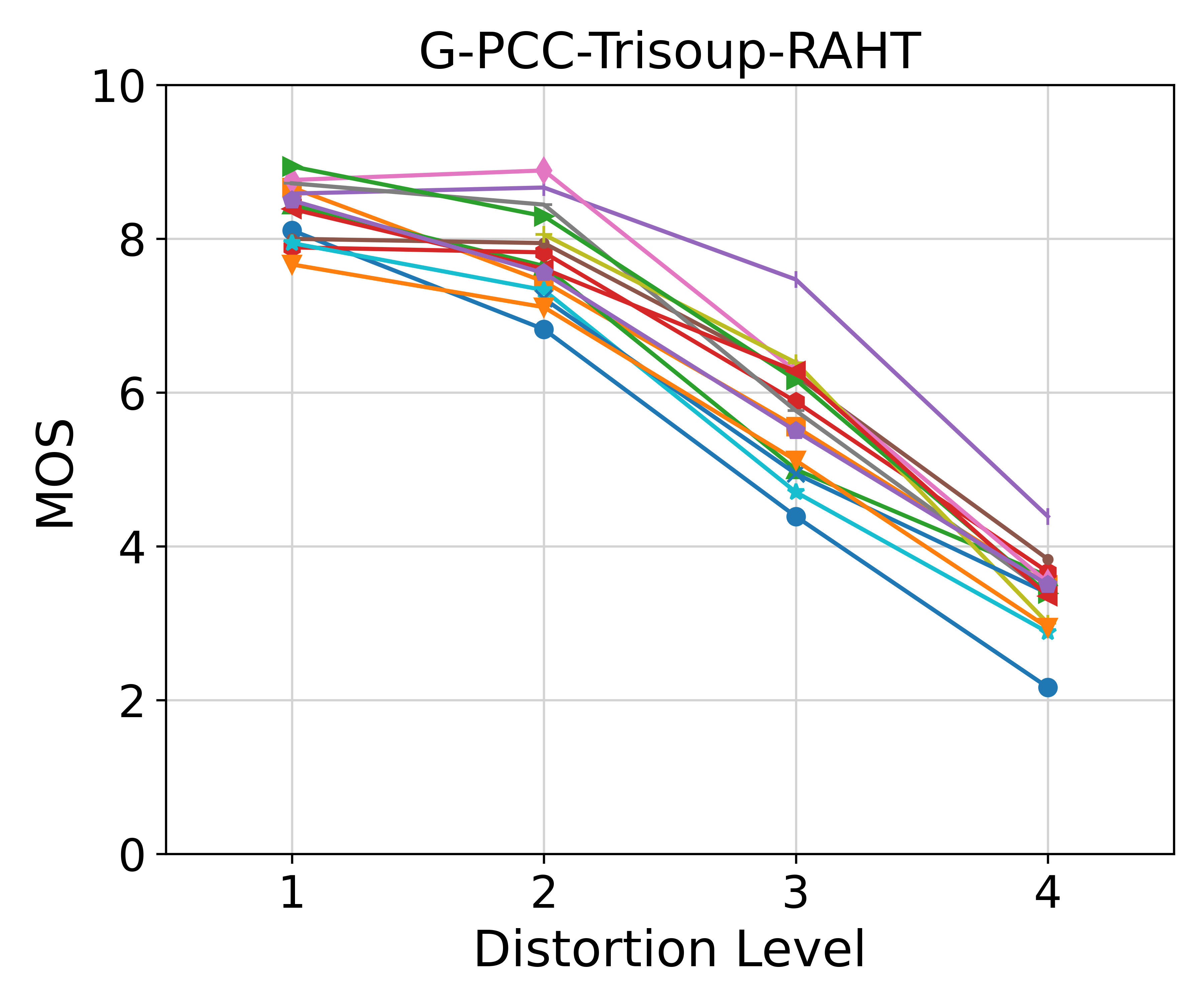}
        \caption{G-PCC-Trisoup-RAHT}
        \label{fig:G-PCC-Trisoup-RAHT}
    \end{subfigure}
    \begin{subfigure}[b]{0.21\textwidth}
        \centering
        \includegraphics[height=2.9cm]{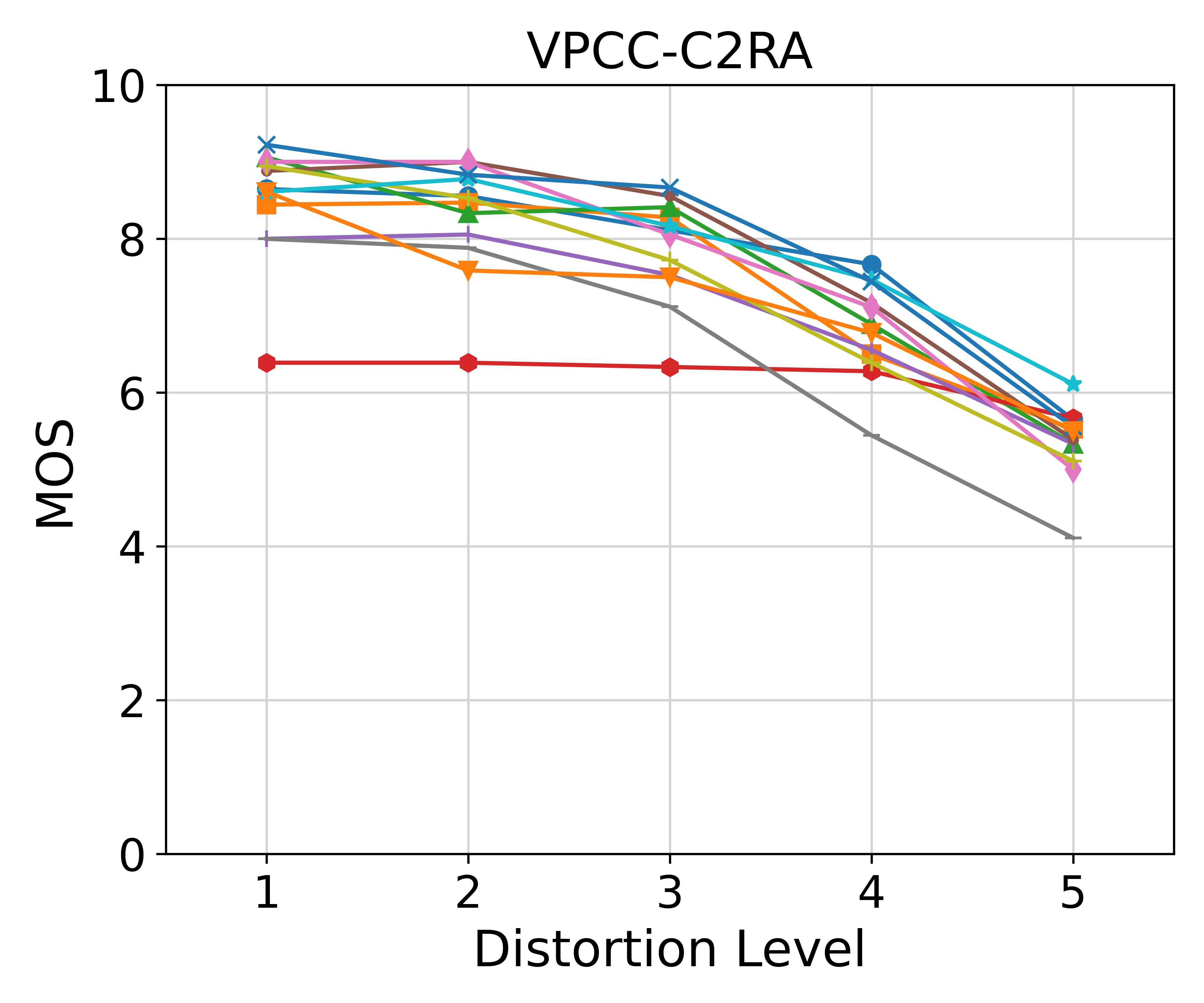}
        \caption{VPCC-C2RA}
        \label{fig:VPCC-C2RA}
    \end{subfigure}
    \begin{subfigure}[b]{0.21\textwidth}
        \centering
        \includegraphics[height=2.9cm]{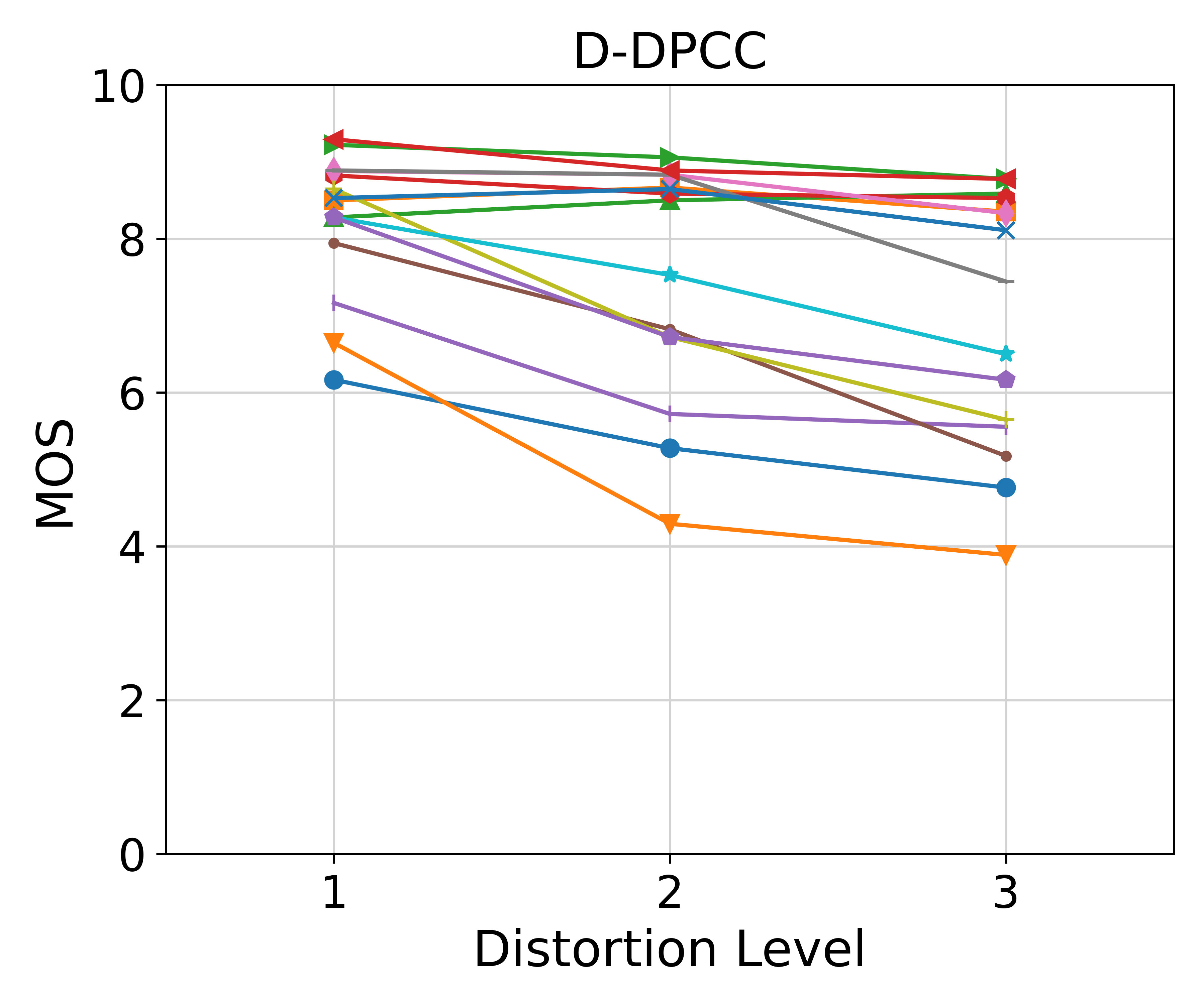}
        \caption{D-DPCC}
        \label{fig:D-DPCC}
    \end{subfigure} 
    \begin{subfigure}[b]{0.21\textwidth}
        \centering
        \includegraphics[height=2.9cm]{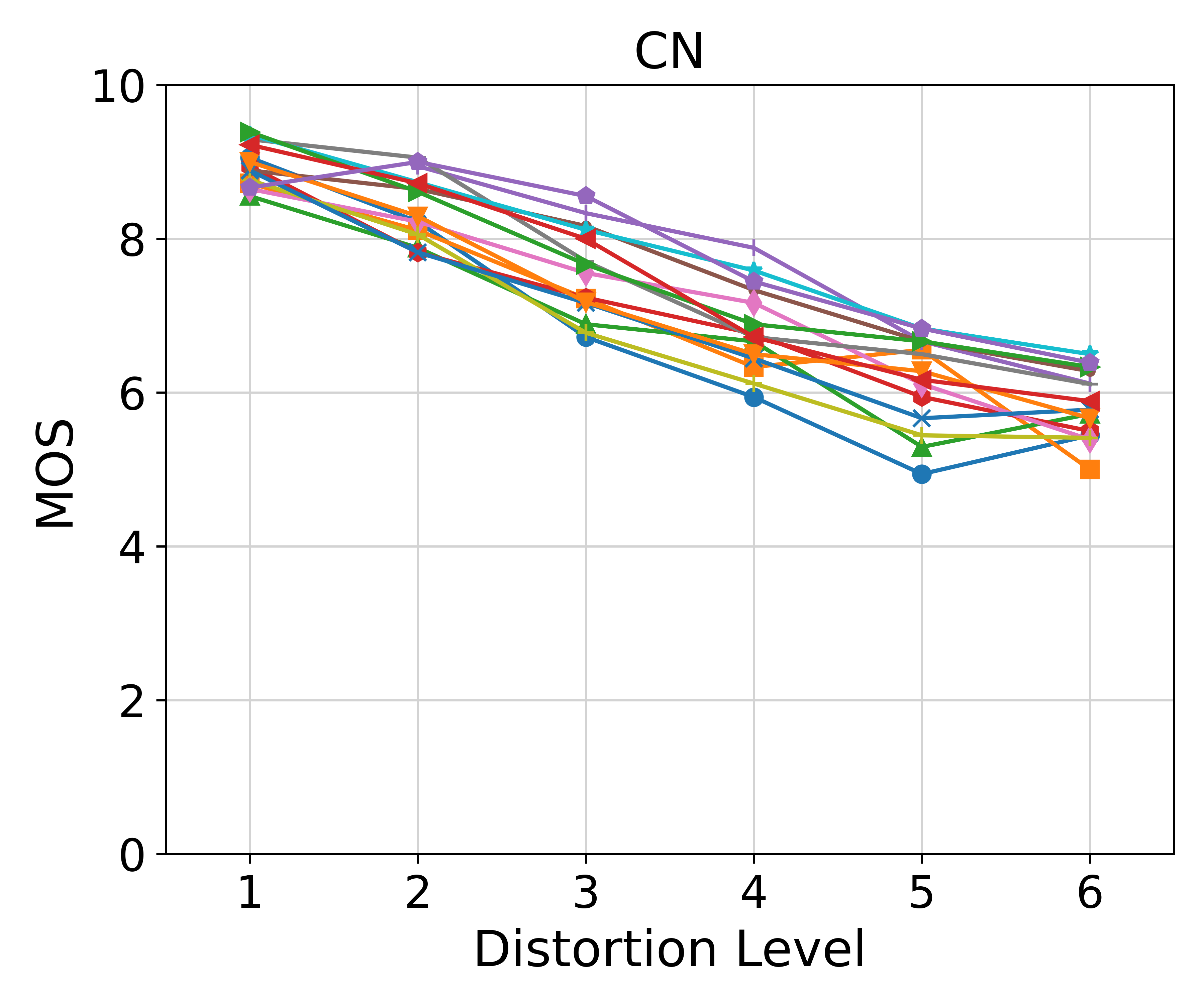}
        \caption{CN}
        \label{fig:CN}
    \end{subfigure}
    \begin{subfigure}[b]{0.21\textwidth}
        \centering
        \includegraphics[height=2.9cm]{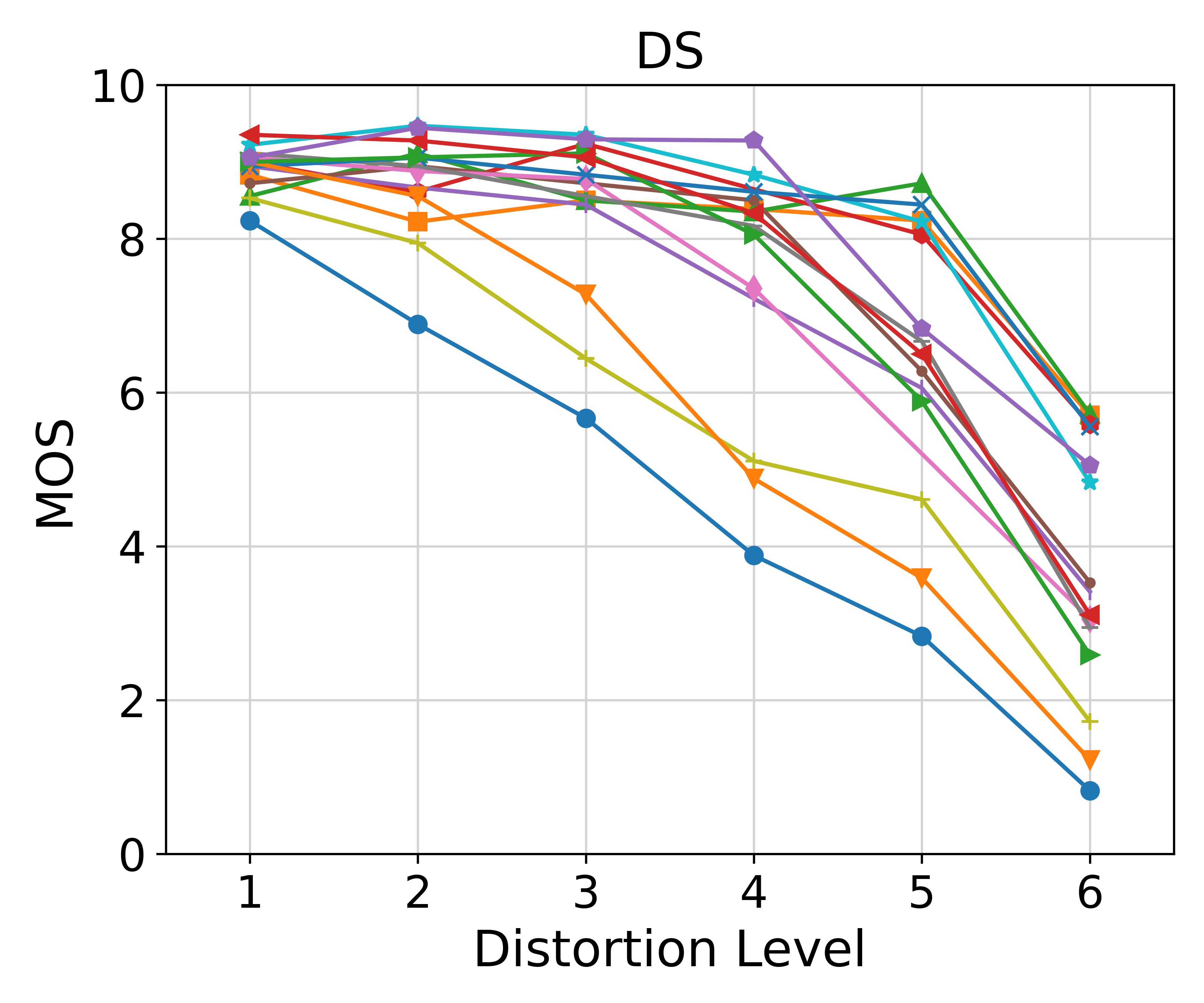}
        \caption{DS}
        \label{fig:DS}
    \end{subfigure}
    \begin{subfigure}[b]{0.21\textwidth}
        \centering
        \includegraphics[height=2.9cm]{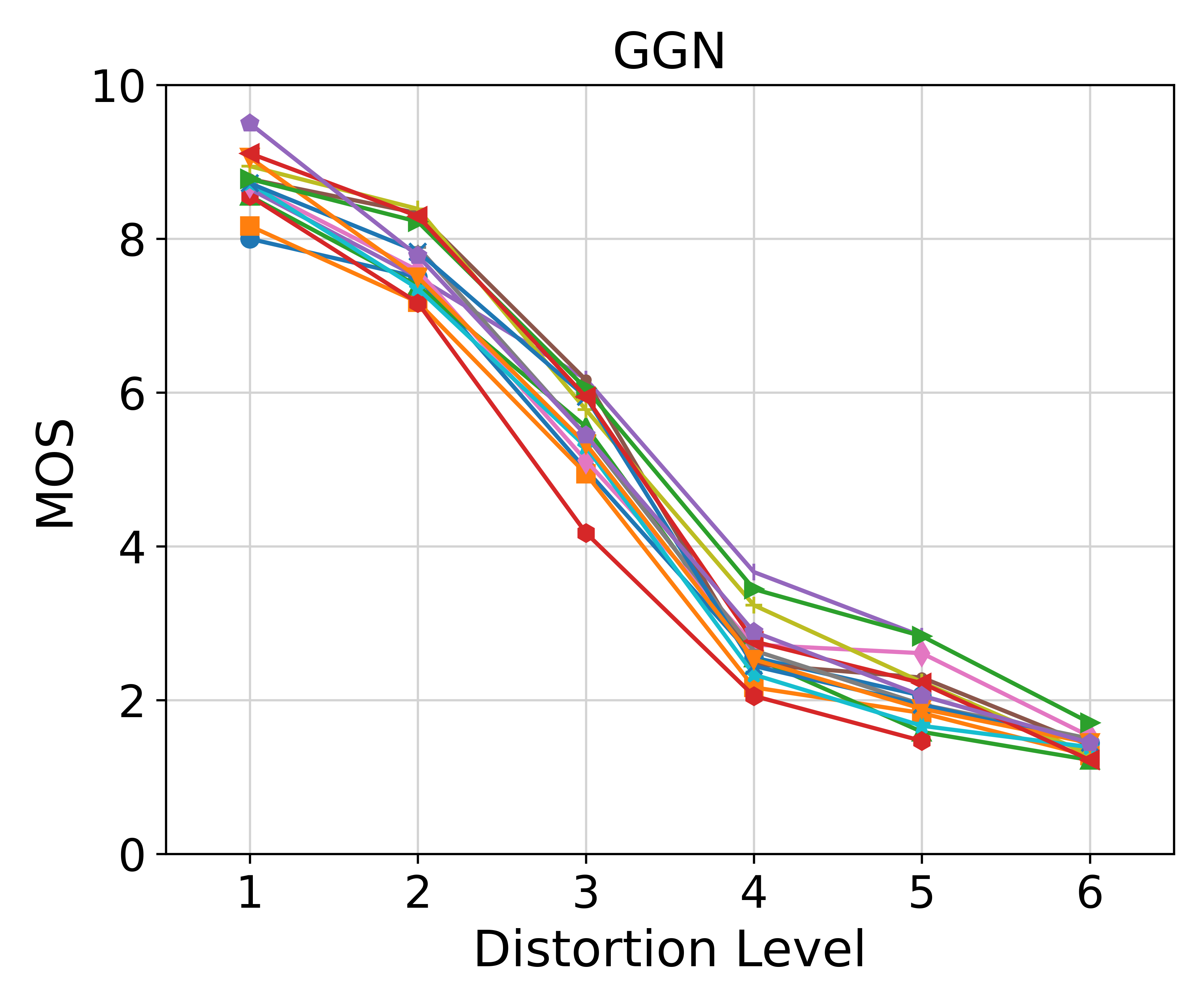}
        \caption{GGN}
        \label{fig:GGN}
    \end{subfigure}
    \begin{subfigure}[b]{0.21\textwidth}
        \centering
        \includegraphics[height=2.9cm]{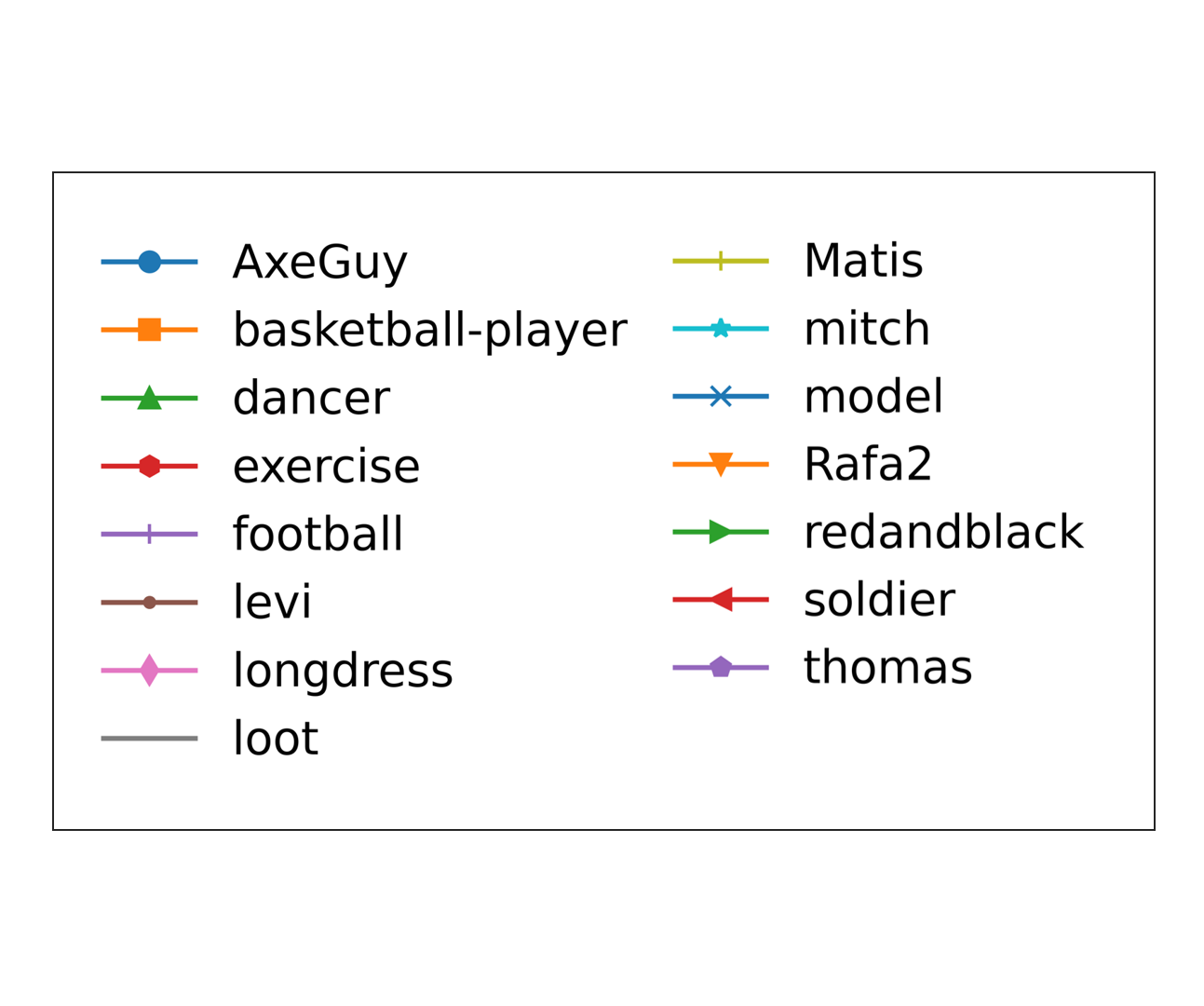}
        \caption{Legend}
        \label{fig:legend}
    \end{subfigure}
\captionsetup{font=normal}
\caption{The MOS distribution under different distortion types.}
\label{fig:mos_vs_distortion}
\end{figure*}

\subsection{MOS Analysis}
We present the MOS distribution in~\Cref{fig:distribution}, where for each score segment, DPCD has at least 50 samples, indicating that the proposed database covers a wide range of quality scores. It is worth noting that the overall MOS of our database is relatively high, with the majority of scores falling within the range of 6 to 10. This phenomenon may be attributed to several factors. On the one hand, the dynamic properties of the samples may mask the distortions, making them less noticeable. On the other hand, the vividness and realism of the human samples may draw participants' attention primarily to the human themselves or their motion, rather than the distortion details.

To validate the accuracy of the MOSs and analyze the impact of different distortions on subjective perception, line graphs of MOSs against distortion types and levels are shown in~\Cref{fig:mos_vs_distortion}. And from these graphs, we can draw the following conclusions:
\textbf{1)} Each graph reveals a general negative correlation between MOS and distortion level, where individual reversals do not affect the overall trend.
\textbf{2)} The lowest score range for Trisoup-RAHT distortion is higher than that for Octree-RAHT. This is because at higher distortion levels, Octree-RAHT significantly decreases the point number and leads to severe information loss, whereas Trisoup-RAHT retains basic geometry and texture information, albeit with distorted quality.
\textbf{3)} The distortion introduced by V-PCC has a relatively small impact on perceived quality. It is worth noting that, the movement of the ``football" cause some points to exceed the coordinate range, resulting in unnatural appearance and lower MOSs. 
\textbf{4)} The distortion introduced by D-DPCC compression is limited. In our experiments, we found that when the lambda is reduced to around 0.1, the bitrate stabilizes and no longer decreases, indicating that this is the maximum achievable distortion. And at the highest distortion level, the MOSs for all samples remain relatively high, which shows the potential of learning-based DPC compression methods. 
\textbf{5)} Samples disturbed by CN provide higher MOSs. This is mainly because CN does not deform the geometry structure of point clouds, and complex textures usually mask tiny noise \cite{zhang2024perception}.
\textbf{6)} For DS, samples with fewer points than other references (\textit{e.g.,} “AxeGuy,” “Matis,” and “Rafa2”) exhibit a sharp decline in MOS as the distortion level increases. In contrast, slight DS has little impact on the other samples with a sufficiently high original point count, as they remain relatively dense even after downsampling. This states that as point clouds become absolutely sparse, perceptual quality is significantly affected.
\textbf{7)} All the lines of GGN are closely aligned. The MOSs for all samples decrease significantly as the distortion level increase, indicating that the human visual perception is highly sensitive to geometric shifts.

\section{Objective Metrics Testing}
\subsection{Metric Selection and Indicators}
Considering the lack of research on objective DPCQA, we test the performance of existing objective SPCQA metrics on DPCD, which can be primarily divided into three categories: point-based, image-based and video-based metrics.

We select 9 point-based metrics adopted by MPEG, 10 widely used image-based metrics and 1 video-based metric. 
For both point-based and image-based metrics, we average the scores across 300 frames for each DPC.
Three common indicators are employed to quantify the efficiency of the objective metrics: Spearman Rank Correlation Coefficient (SRCC), Pearson Linear Correlation Coefficient (PLCC), and Root Mean Square Error (RMSE). To ensure consistency between the value ranges of the predicted scores and MOSs, a nonlinear four-parameter logistic fitting function is used to align their ranges following~\cite{fittingfunction}.

\begin{table*}[t]\footnotesize
\centering 
\renewcommand\arraystretch{0.9}
\captionsetup{font=normal}
\caption{The performance results of objective metrics. ``P" stands for point-based metrics, ``I" represents image-based metrics and ``V" represents video-based metrics. The symbol `--' indicates that the results of the metric for samples with this kind of distortion are meaningless.
Best and second performance results are marked in {\bf\textcolor{red}{RED}} and {\bf\bf\textcolor{blue}{BLUE}}.}

\begin{tabular}{c|c|c|ccc|ccccccc}
\toprule
    \multirow{2}{*}{Type} & \multirow{2}{*}{Metric} & \multirow{2}{*}{Reference} & \multicolumn{3}{c|}{Overall} & \multicolumn{2}{c}{G-PCC} & V-PCC & \multirow{2}{*}{D-DPCC} & \multirow{2}{*}{CN} & \multirow{2}{*}{DS} & \multirow{2}{*}{GGN} \\
    & & &  SROCC & PLCC & RMSE & Octree & Trisoup & C2RA & & & \\ 
    \hline
    \noalign{\vskip 0.3em} 
\multirow{9}{*}{P} 
 & P2point\_MSE~\cite{mekuria2016evaluation} & \checkmark & \bf\textcolor{blue}{0.891} & 0.902 & 1.185 & \bf\textcolor{blue}{0.925} & \bf\textcolor{blue}{0.876} & 0.849 & 0.709 & -- & 0.806 & 0.892 \\
 & P2point\_MSE\_PSNR~\cite{mekuria2016evaluation} & \checkmark & \bf\textcolor{red}{0.899} & \bf\textcolor{red}{0.921} & \bf\textcolor{blue}{1.073} & \bf\textcolor{red}{0.945} & \bf\textcolor{red}{0.893} & \bf\textcolor{blue}{0.863} & 0.722 & -- & 0.805 & 0.892 \\
 & P2point\_Haus~\cite{mekuria2016evaluation} & \checkmark & 0.532 & 0.546 & 2.303 & 0.840 & 0.807 & 0.659 & 0.571 & -- & 0.758 & 0.921 \\
 & P2point\_Haus\_PSNR~\cite{mekuria2016evaluation} & \checkmark & 0.551 & 0.582 & 2.234 & 0.867 & 0.850 & 0.685 & 0.577 & -- & 0.764 & 0.922 \\
 & P2plane\_MSE~\cite{tian2017geometric} & \checkmark & 0.778 & 0.745 & 1.833 & 0.808 & 0.752 & 0.856 & \bf\textcolor{red}{0.806} & -- & 0.756 & 0.778 \\
 & P2plane\_MSE\_PSNR~\cite{tian2017geometric}  & \checkmark & 0.780 & 0.771 & 1.751 & 0.806 & 0.746 & \bf\textcolor{red}{0.874} & \bf\textcolor{blue}{0.795} & -- & 0.755 & 0.776 \\
 & P2plane\_Haus~\cite{tian2017geometric} & \checkmark & 0.627 & 0.611 & 2.176 & 0.765 & 0.766 & 0.652 & 0.768 & -- & 0.761 & 0.839 \\ 
 & P2plane\_Haus\_PSNR~\cite{tian2017geometric}& \checkmark  & 0.634 & 0.626 & 2.143 & 0.779 & 0.773 & 0.653 & 0.764 & -- & 0.763 & 0.838 \\ 
 & PSNR\_yuv & \checkmark & 0.705 & 0.657 & 1.946 & 0.904 & 0.714 & 0.726 & 0.693 & \bf\textcolor{blue}{0.927} & 0.618 & 0.778 \\
 \hdashline
 \noalign{\vskip 0.3em} 
\multirow{10}{*}{I} 
 & PSNR & \checkmark & 0.717 & 0.728 & 1.760 & 0.664 & 0.508 & 0.637 & 0.576 & 0.878 & 0.838 & 0.916 \\
 & SSIM & \checkmark & 0.736 & 0.738 & 1.732 & 0.632 & 0.505 & 0.691 & 0.603 & 0.756 & 0.763 & 0.923 \\
 & MS-SSIM~\cite{wang2003multiscale} & \checkmark & 0.735 & 0.735 & 1.742 & 0.634 & 0.507 & 0.677 & 0.594 & 0.760 & 0.762 & 0.910 \\
 & IW-SSIM~\cite{wang2010information} & \checkmark & 0.713 & 0.721 & 1.779 & 0.635 & 0.508 & 0.681 & 0.534 & 0.796 & 0.747 & 0.929 \\
 & VIF~\cite{sheikh2006image} & \checkmark & 0.694 & 0.719 & 1.785 & 0.652 & 0.546 & 0.715 & 0.489 & 0.867 & 0.816 & \bf\textcolor{blue}{0.948} \\
 & FSIM~\cite{zhang2011fsim} & \checkmark & 0.698 & 0.715 & 1.797 & 0.616 & 0.503 & 0.681 & 0.590 & 0.778 & 0.777 & 0.904 \\ 
 & LPIPS~\cite{zhang2018unreasonable} & \checkmark & 0.795 & 0.776 & 1.620 & 0.671 & 0.567 & 0.783 & 0.722 & 0.883 & \bf\textcolor{blue}{0.851} & 0.927 \\
 & DISTS~\cite{ding2020image} & \checkmark & 0.887 & \bf\textcolor{blue}{0.910} & \bf\textcolor{red}{1.066} & 0.900 & 0.751 & 0.812 & 0.721 & \bf\textcolor{red}{0.929} & \bf\textcolor{red}{0.879} & \bf\textcolor{red}{0.955} \\
 & CLIP-IQA~\cite{wang2023exploring} & $\times$ & 0.275 & 0.523 & 2.189 & 0.622 & 0.313 & 0.299 & 0.065 & 0.269 & 0.600 & 0.516 \\
 & BRISQUE~\cite{mittal2012no} & $\times$ & 0.572 & 0.601 & 2.053 & 0.822 & 0.553 & 0.622 & 0.311 & 0.629 & 0.606 & 0.575 \\
 \hdashline
 \noalign{\vskip 0.3em} 
\multirow{1}{*}{V}
 & VMAF~\cite{vmaf} & \checkmark & 0.650 & 0.676 & 1.893 & 0.758 & 0.556 & 0.652 & 0.465 & 0.699 & 0.725 & \bf\textcolor{blue}{0.948} \\
\bottomrule
\end{tabular}
\label{tab:experiment}
\end{table*}

\subsection{Overall Performance}
The performance of the metrics on the entire database are shown in the ``Overall” columns of~\Cref{tab:experiment}. Based on these results, the following conclusions can be drawn:
\textbf{1)} Among the point-based metrics, the two MSE-based P2point approaches yield the best. In comparison, P2plane underperforms, likely due to the errors introduced during estimation of normal vectors. Additionally, normalizing the computation results using bounding boxes and converting them to the corresponding PSNR values improves performance by standardizing the scale. 
\textbf{2)} Among the image-based metrics, DISTS and LPIPS achieve the highest performance. By leveraging networks pre-trained on large-scale image datasets, these metrics effectively capture representative features, thereby enhancing their generalizability.
\textbf{3)} The video-based metric VMAF, despite considering temporal information, does not yield superior results. This may be because VMAF primarily focuses on temporal variations in natural scenes, while our database comprises individual human point cloud samples.
\textbf{4)} Despite the inherent information loss in image-based metrics, their performance can rival that of point-based metrics. This can be attributed primarily to the fact that image-based metrics excel at extracting texture information, while point-based metrics tend to focus more on geometry and may not fully exploit multimodal data.
\textbf{5)} All the no-reference metrics report noticeably poorer performance compared to full-reference metrics. The lack of reference samples as a benchmark prevents accurate assessment of distortions, thus limiting the evaluation accuracy.

\subsection{Analysis by Type of Distortion}
For a more comprehensive analysis, we further provide the SRCC results for different types of distortion in~\Cref{tab:experiment}. The following conclusions can be derived from these results:
\textbf{1)} The two MSE-based P2point approaches demonstrate the best performance on G-PCC. Since G-PCC typically introduces geometric distortions, P2point metrics, which directly measure the Euclidean distance between corresponding points in the distorted and reference point clouds, are more sensitive to such distortions.
\textbf{2)} P2plane\_MSE\_PSNR performs the best on V-PCC, while P2plane\_MSE performs the best on D-DPCC. MSE-based metrics outperform Hausdorff distance-based metrics, as the latter involve maximum pooling, which may cause outliers with large coordinate values in the point cloud to negatively impact the final result. 
\textbf{3)} DISTS demonstrates robustness across various distortions and achieves the best results on CN, DS, and GGN, with SRCC values of approximately 0.929, 0.879, and 0.955, respectively, due to its ability to effectively capture both local and global information.

\subsection{Weakness of Current Metrics}
Overall, current metrics exhibit several limitations, which are summarized as follows:
\textbf{1)} For point-based metrics, while MSE-based P2point metrics perform well, they still have room for improvement. Additionally, the high computational complexity makes them impractical for real-world applications. 
\textbf{2)} Image and video-based metrics may suffer from information loss during projection, potentially masking original distortions. Moreover, their performance can be influenced by background information, leading to unstable scores across different contents.
\textbf{3)} No approach consistently performs well across all distortion types. 
Specifically, while P2point is sensitive to traditional compression, it struggles with measuring color distortions. LPIPS and DISTS are effective for CN but perform poorly on traditional compression methods. Moreover, most metrics exhibit inferior performance on the learning-based DPC compression.
Traditional point-based metrics, as well as existing image-based and video-based metrics, may overlook the unique characteristics and distortions of DPCs, leading to inaccurate quality prediction on specified distortions. Therefore, there is a strong need for effective objective metrics specifically tailored to DPCs. And our proposed database may facilitate the design of such metrics.

\section{Conclusion}
In this paper, we create a large-scale dynamic point cloud database DPCD, which consists of 15 reference DPCs and 511 distorted samples with accurate MOSs. The database undergoes comprehensive
analysis to validate its content diversity, illustrate the characteristics of different distortion types, and assess its MOS accuracy. Additionally, we evaluate several commonly used objective metrics on DPCD. The best full-reference metrics achieve a correlation around 0.90, while all the no-reference metrics are struggle with DPC quality prediction with only 0.28 to 0.57 correlations. With the accurate and large-scale MOS labels, our database can serve as a benchmark for objective metrics, and further promote the algorithms related to DPCs in the future.

\section*{Acknowledgment}
This paper is supported in part by National Natural Science Foundation of China (62371290), National Key R$\&$D Program of China (2024YFB2907204), the Fundamental Research Funds for the Central Universities of China, and STCSM under Grant (22DZ2229005). The corresponding author is Yiling Xu (e-mail: yl.xu@sjtu.edu.cn). 

\bibliographystyle{IEEEbib}
\bibliography{icme2025}
\end{document}